\documentclass[times,twocolumn,final]{article}
\usepackage{authblk}
\usepackage{framed,multirow}
\usepackage{amssymb}
\usepackage{latexsym}
\usepackage{amsmath}
\usepackage{url}
\usepackage{xcolor}
\usepackage{hyperref}
\usepackage{graphicx}

\definecolor{newcolor}{rgb}{.8,.349,.1}
\newcommand*\samethanks[1][\value{footnote}]{\footnotemark[#1]}
\providecommand{\keywords}[1]{\textbf{\textit{\\Keywords---}} #1}
\date{}
\begin{document}

\title{Understanding the effects of artifacts on automated polyp detection and incorporating that knowledge via learning without forgetting}

\author[2]{Maxime Kayser \thanks{M. K. was with Technische Universit\"{a}t M\"{u}nchen, Munich, Germany at the moment of this research.}\thanks{M. K. and R. D. S. share first authorship.}}

\author[1]{Roger D. Soberanis-Mukul \samethanks[2]\thanks{Corresponding author: roger.soberanis@tum.de}}

\author[5]{Anna-Maria Zvereva M.D.}
\author[6]{Peter Klare M.D. \thanks{P. K. was with Klinikum rechts der Isar der Technischen Universit\"{a}t M\"{u}nchen, Munich, Germany at the moment of this research}}

\author[1,3]{Nassir Navab\thanks{N. N. and S. A. share senior authorship.}}
\author[1,4]{Shadi Albarqouni\samethanks[5]}

\affil[1]{Computer Aided Medical Procedures, Technische Universit\"{a}t M\"{u}nchen, Munich, Germany}
\affil[2]{Big Data Institute, University of Oxford, Oxford, UK}
\affil[3]{Computer Aided Medical Procedures, Johns Hopkins University, Baltimore, USA}
\affil[4]{Computer Vision Laboratory, ETH Zurich, Zurich, Switzerland}
\affil[5]{Klinik f\"{u}r Innere Medizin II, Klinikum rechts der Isar der Technischen Universit\"{a}t M\"{u}nchen, Munich, Germany}
\affil[6]{Abteilung Innere Medizin Gastroenterologie, Krankenhaus Agatharied Hausham, Germany}

\maketitle
\begin{abstract}
Survival rates for colorectal cancer are higher when polyps are detected at an early stage and can be removed before they develop into malignant tumors. Automated polyp detection, which is dominated by deep learning based methods, seeks to improve early detection of polyps. However, current efforts rely heavily on the size and quality of the training datasets. The quality of these datasets often suffers from various image artifacts that affect the visibility and hence, the detection rate. In this work, we conducted a systematic analysis to gain a better understanding of how artifacts affect automated polyp detection. We look at how six different artifact classes, and their location in an image, affect the performance of a RetinaNet based polyp detection model. We found that, depending on the artifact class, they can either benefit or harm the polyp detector. For instance, bubbles are often misclassified as polyps, while specular reflections inside of a polyp region can improve detection capabilities. We then investigated different strategies, such as a learning without forgetting framework, to leverage artifact knowledge to improve automated polyp detection. Our results show that such models can mitigate some of the harmful effects of artifacts, but require more work to significantly improve polyp detection capabilities.\end{abstract}
\keywords{Polyp detection, Artifact detection, Learning without forgetting, Multi-task learning}

\section{Introduction} \label{sec:introduction}
\begin{figure}
\begin{center}
\includegraphics[width=0.95\columnwidth]{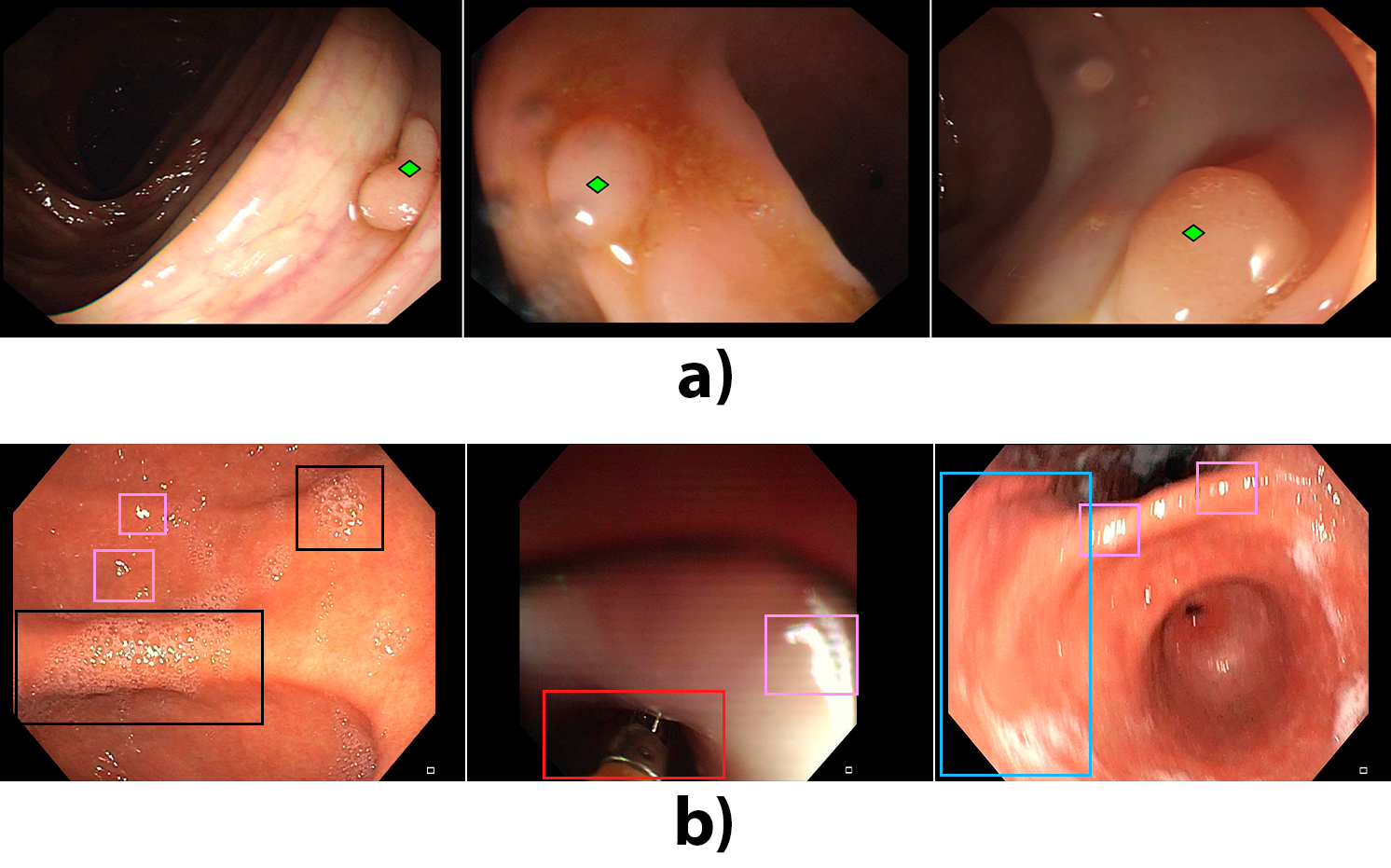}
\end{center}
\caption{a) Sample frames from the CVC-ClinicDB dataset. Polyps locations are indicated by green marks. b) Samples from the EAD challenge dataset. Shown artifact classes are specularity (pink box), instruments (red), bubbles (black), and blur (blue). Additional artifact classes include contrast, saturation, and misc. artifact, which aggregates all other types of artifacts.} \label{fig:cvc_samples}
\end{figure}

    Colorectal cancer (CRC) is the most deadly cancer in the United States \cite{siegel2018cancer}. CRC often develops from polyps, which may turn into tumors (Fig. \ref{fig:cvc_samples}a). Early detection of these polyps can increase the CRC survival rate to up to 95\% \cite{bib:bernal2012}. The standard procedure for polyp screening is the endoscopic analysis of the colon, called colonoscopy. One of the ways to reduce the polyp miss rate, which is crucial for minimizing the risks related to CRC, is the use of a real-time automated polyp-detection system. However, automated polyp detection remains a challenging problem, given the high variation in appearance, size, and shape of polyps, their often similar texture to the surrounding tissue, as well as image artifacts obstructing and corrupting the endoscopic video frames. 

Initial work \cite{bernal2017comparative} has shown how some artifacts, such as specularity (i.e. strong reflection of light) or blur, can have a significant effect on the performance of automated polyp detection systems. However, this study, as it required manual labelling of artifacts, is limited to a restricted set of artifact classes and labels them on an image level, without specifying their location. We extended their work on understanding how artifacts affect automated polyp detection performance by considering an inclusive set of six different artifacts and by specifying their location via bounding boxes. This allowed us to specifically analyse how our deep learning based polyp detection model is affected when looking at regions that overlap with artifacts or have artifacts inside of them. We annotated the artifacts with a model that has been trained on an endoscopic artifact dataset released by the Endoscopic Artefact Detection (EAD) challenge \cite{DBLP:journals/corr/abs-1905-03209} and has ranked third place in that challenge.

While previous literature has suggested that artifacts have a significant influence on automated polyp detection, no work has yet tried to incorporate knowledge about artifacts in deep learning based polyp detection models. Existing work either attempted to restore the frames, which is currently not feasible in real-time \cite{ali2019deep} or approached polyp and artifact detection as a simple multi-class detection problem \cite{DBLP:journals/corr/VazquezBSFLRDC16}. In this work, we addressed this knowledge gap by exploring ways to incorporate artifact information in polyp detection. The hypothesis is that, by teaching a model to represent artifacts, its ability to distinguish between artifacts and polyps will be improved. Amongst others, we built a RetinaNet based model that is modified according to the learning without forgetting (LwF) \cite{8107520} approach. LwF aims to make a model learn new capabilities while maintaining performance on the old capabilities, without relying on the training data from the old tasks. It has been shown that LwF can thereby improve performance on the new task \cite{8107520}.

Our contributions in this work are thus twofold: 1) we present the most extensive existing analysis of how artifacts affect polyp detection and 2) we present the first work that explores multi-task learning (MTL) techniques for including artifact information in deep learning based polyp detection.

\section{Related work} \label{sec:related_work}
    In this section, we looked at related literature in polyp detection, artifact detection, and multi-task learning.

\subsection{Polyp Detection} \label{subsec:polyp_detection}
Automated polyp detection is a computer vision problem that has been dominated by hand-crafted features methods before computing power has made deep learning possible \cite{karkanis2003computer,iakovidis2005comparative,alexandre2008color,ameling2009texture,gross2009comparison,hwang2007polyp,ganz2012automatic,BERNAL201599}. However, given the great variance of polyp shapes, different angles used in colonoscopy, and the different lighting modes applied, the appearance of polyps varies widely and therefore these methods have always had limited performance \cite{yu2016integrating}. With the ascent of deep learning, automated polyp detection performance has improved significantly in recent years and the majority of high-performing approaches now rely on it \cite{bernal2017comparative,zhu2015lesion,yu2016integrating,brandao2017fully,angermann2017towards,wang2018development,mohammed2018net,shin2018automatic,brandao2018towards,bib:ximo18,bib:kang19,bib:jia19}. Some methods adopt a hybrid approach, where hand-crafted and learned features are combined \cite{tajbakhsh2015automated,bae2015polyp,silva2014toward,vsevo2016edge}. Recent work includes \cite{bib:ximo18}, where they show that Faster R-CNN \cite{ren2015faster}, a popular object detection framework that relies on a region proposal module, can achieve state-of-the-art performance. \cite{bib:jia19} proposed a two-step polyp detection framework, which uses Faster R-CNN with feature pyramid networks to detect polyp regions and segments them with a fully convolutional network in the second step. In another polyp segmentation approach, \cite{bib:kang19} proposed an ensemble of two Mask R-CNN models \cite{he2017mask} with different backbone architectures. However, so far none of these works has tried to leverage artifact information in the polyp detection frameworks.

\subsection{Artifact Detection} \label{subsec:artifact_detection}
A study on MICCAI’s 2015 polyp detection competition has shown that automated polyp detection is still hampered by the artifacts contained in endoscopic frames \cite{bernal2017comparative}. In that study, three experts were asked to label the ASU-Mayo Clinic Colonoscopy Video Database \cite{tajbakhsh2015automated} for massive specular highlights presence, low visibility, specular highlights within polyp, and overexposed regions. They then looked at how the performance of the different models participating in the competition differs in the presence of these artifacts. We were able to extend the findings of this study by involving a higher number of artifacts (Fig. \ref{fig:cvc_samples}b) and, by having artifact annotations at a bounding box and not at an image label level, conduct a more fine-grained analysis of the effects that artifacts have on polyp detection algorithms.

Efforts have also been made to automatically detect these artifacts in order to remove them and restore the images. Most of these approaches focus only on specific artifacts, such as  blur or specular reflections \cite{stehle2006removal, tchoulack2008video, liu2011blurring, funke2018generative, akbari2018adaptive}. A recent work has built a framework that can detect six different artifact classes and apply artifact-specific frame restoration procedures, which are often based on adversarial networks \cite{ali2019deep}. While this method can restore the quality of endoscopic frames for post-processing procedures, it is not applicable in real-time. In contrast to these efforts, we seek to improve polyp detection in real-time, by teaching a model to differentiate between artifacts and polyps.  

To our best knowledge, the only work that has addressed artefacts in the context of polyp detection is \cite{vazquez2017benchmark}. This team has extended an existing polyp database by adding segmentation annotation for lumen, specularity, borders, and background. Their aim was to build a model that is able to detect and segment polyps as well as the four above mentioned classes. In contrast to our work, the researchers did not aim to use artifact data to improve polyp detection. In fact, training their model to be able to detect all five classes has significant detrimental effects on their polyp detection performance. Compared to them, we included more artifact classes and investigated more sophisticated approaches to leverage artifacts for automated polyp detection.

\subsection{Multi-task Learning} \label{subsec:mtl}
In our context, we will refer to multi-task learning (MTL) as the approaches that use auxiliary (or related) tasks to improve the original primary task. Early work has shown that MTL can for example improve mortality rate prediction of pneumonia patients \cite{caruana1997multitask}.  The main inputs for this model are certain patient characteristics, such as age or whether the patient presents determined symptoms. By ensuring that the model not only predicts mortality rate but also related outputs such as white blood cell count, the hidden layers of the model were biased to better capture certain characteristics of the patient, which lead to better model performance. More recent work presented by \cite{zhou19} used the correlation between the severity of diabetic retinopathy and lesions present on the eye. The authors proposed a model for disease grading and multi-lesion segmentation that allows both tasks to collaborate to improve each other. Their proposal also permits the segmentation model to train in a semi-supervised way. This is done by using the segmenter's predictions as input for an attentive model used by the grading network. At the same time, the attention maps generated by this attentive model are used as pseudo-mask for training the segmenter with unannotated images. In our work, the primary task is polyp detection and the auxiliary task is artefact detection.

\section{Methodology} \label{sec:methodology}
    In this section we describe the single-task polyp detection and artifact detection models, which were used to analyse the effect that artifacts have on polyp detection. We also describe our proposed multi-task model, which uses LwF to leverage artifact information to improve polyp detection.

\subsection{Problem Formulation} \label{subec:problem_formulation}

Given an input frame $X \in \mathbb{R}^{h\times w \times 3}$  from a colonoscopy video sequence, we defined single-task models as $Y^t = f(X;\theta ^t)$, where $\theta ^t$ are the model parameters trained for a specific task $t$. Thus, our polyp detection model is given by $Y^p = f(X;\theta ^p)$ and our artifact detection model is given by $Y^a = f(X;\theta ^a)$. Both single-task models are based on RetinaNet, which is described in section \ref{subsec:retinanet}. 

The polyp detector outputs a set of bounding boxes $Y^p = \{y_i ^p; i = 1,\dots n_p\}$ where $y_i ^p$ represents a bounding box detection in $X$  and $n_p$ is the total number of detections. It trains on a polyp dataset $D^p = \{(X_i ^p, Y_i ^p); i = 1, \dots, m_p \}$, where $m_p$ is the number of samples. 

The artifact detector gives us an equivalent set of artifact locations $Y^a$, and an additional set $C = \{c_i; i = 1, \dots, n_c\}$ of artifact classes $c_i \in \{0,\dots, 5\}$, with $n_c$ the total number of artifacts found in $X$. Here $c_i$ indicates the type of artifact in the box $y_i ^a$. Artifact types are blur, bubbles, contrast, specularity, saturation, and miscellaneous (misc.) artifacts. The dataset for training this detector is defined as $D^a = \{(X_i ^a, Y_i ^a, C_i); i = 1, \dots, m_a \}$, where $m_a$ is the number of samples.

For the MTL model, our main objective was to improve polyp detection performance by including information about the artifacts present in $X$. We thus defined the MTL model as $Y^t = f(X; \theta ^s, \theta ^p)$, where $\theta ^s$ is a set of shared parameters across artefact and polyp detection tasks and $\theta ^p$ represents polyp detection specific parameters. This model is inspired by LwF, which is explained in section \ref{subsec:lwf}.

\subsection{RetinaNet Base Model} \label{subsec:retinanet}

RetinaNet \cite{lin2017focal} consists of a backbone network for extracting convolutional feature maps and two subnetworks that perform object classification (classification subnetwork) and bounding box regression (regression subnetwork) via convolutional layers. The backbone network can consist of conventional image classification convolutional neural network (CNN), such as ResNet \cite{he2016deep}. The classification and regression losses are given by the focal loss and the smooth L1 loss, respectively. It is a one-stage method, meaning that it does not require a region proposal module as is common in object detection \cite{girshick2014rich, girshick2015fast, ren2015faster}. Instead, anchors at different scales and aspect ratios are densely distributed across the image and classified by the network. To construct a multi-scale feature pyramid from a single resolution input image, the backbone network is augmented by a feature pyramid network (FPN) \cite{lin2017feature}.

The main innovation of RetinaNet is focal loss. It is a modification of the cross-entropy loss that adds weighting parameters to avoid one-stage detectors from being swamped by the great number of easy background (i.e. non object) anchors. To address this imbalance, focal loss introduces a weighting factor of $(1-q^{\mathrm{*}})^\gamma$, where $q^{\mathrm{*}}$:

\begin{equation} \label{eq:pt}
q^{\mathrm{*}}=\left\{\begin{array}{ll}{q} & {\hbox { if } y=1} \\ {1-q} & {\hbox { otherwise, }}\end{array}\right.
\end{equation}

where $q$ is the estimated probability $P(y=1)$ of a detection. This factor reduces the importance of easy anchors. $\gamma$ is a hyperparameter that controls the extent to which the loss focuses on hard examples. For $\gamma = 0$, the focal loss is the same as the cross-entropy loss. Focal loss incorporates another weighting factor, $\alpha \in [0,1]$  to address the class imbalance between background and foreground anchors. Foreground anchors will be weighted by $\alpha$ and background anchors will be weighted by $1 - \alpha$. We define $\alpha^{\mathrm{*}}$ analogously to $q^{\mathrm{*}}$ in equation \ref{eq:pt}. Then, the focal loss is given by: 

\begin{equation} \label{eq:fl}
\mathrm{FL}\left(q^{\mathrm{*}}\right)=-\alpha^{\mathrm{*}}\left(1-q^{\mathrm{*}}\right)^{\gamma} \log \left(q^{\mathrm{*}}\right).
\end{equation}

\subsection{Single-task Models}  \label{subsec:stl}

Our single-task artifact and polyp detection models are based on RetinaNet.  We use these models to evaluate the impact of artifacts present in the image (see section \ref{subsec:subsec_art_inf}). The base polyp detector $Y^p = f^*(X;\theta ^p)$ uses a standard RetinaNet architecture with a ResNet that was pre-trained on ImageNet as backbone network. The base artifact detector $[Y^a, C] = f^*(X;\theta ^a)$ is taken from our submission to  EAD 2019 challenge \cite{kayser2019focal}. For the competition, we validated the model for optimal classification/regression loss weighting,  focal  loss  parameters, data augmentation,  and  backbone  model configurations (ResNet  50, 101, and 152). In the competition we used an ensemble of seven models and ranked third overall. In this work we did not use the ensemble model, but our best performing single model from the challenge.

\subsection{Learning Without Forgetting} \label{subsec:lwf}

Learning without Forgetting (LwF) is an MTL strategy that allows to teach an additional task $t^p$ to an existing model previously trained on a related task $t^a$ \cite{8107520}. One of the main advantages of LwF is that, to extend the capabilities of a model, only the training data of the new task is necessary. In our case, all we needed was thus an initial artifact detector $[Y^a, C] = f(X, \theta^s, \theta^a)$ with $\theta^a$ initial task-specific parameters, and a dataset $D^p$ for our new polyp detection task. 

\begin{figure*}[ht]
    \centering
    \includegraphics[width=\textwidth]{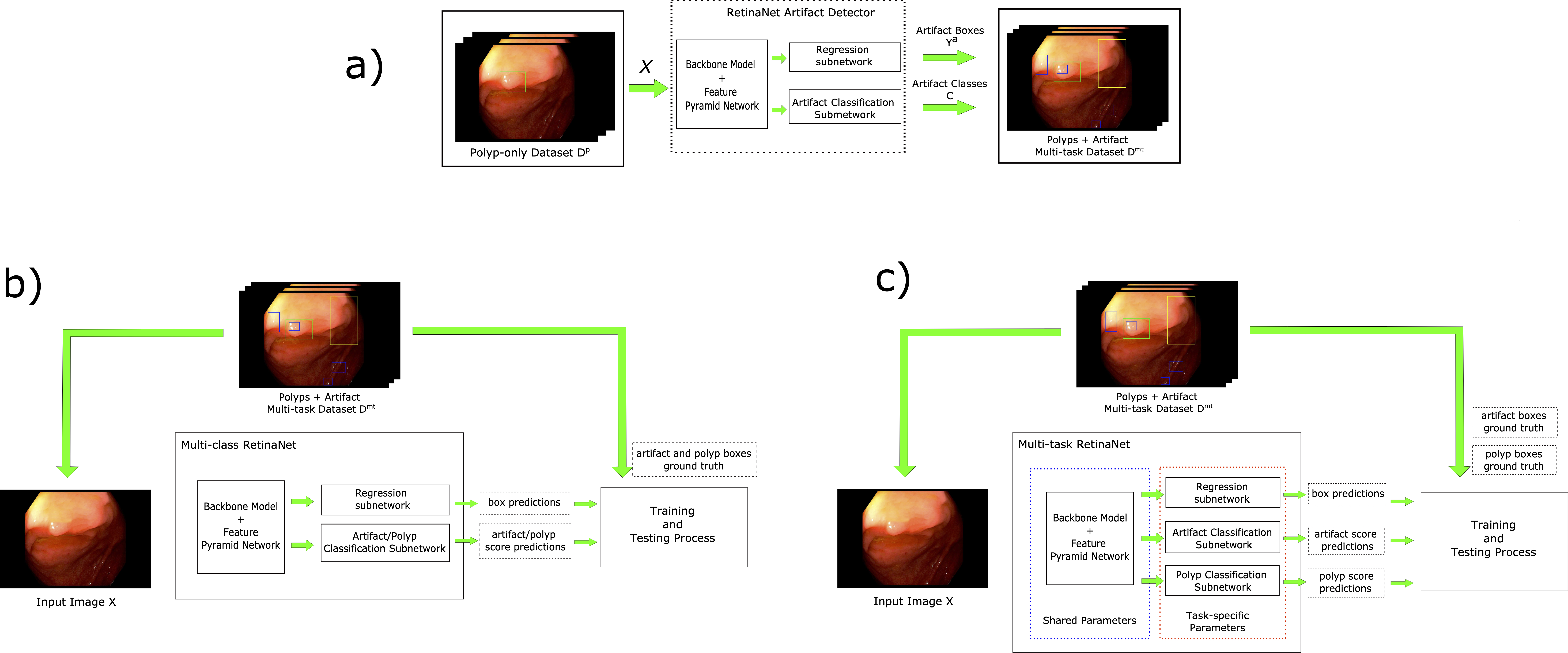}
    \caption{a) Dataset $D^{mt}$ is generated by using our artifact detection model to annotate artifact bounding boxes in a polyp-only dataset. b) This depicts our approach for modeling the problem as a multi-class object detection problem. The model does not differentiate between polyps and different artifact classes. c) This depicts our learning without forgetting (LwF) inspired approach. The model has a set of shared parameters (backbone model, feature pyramid network, and regression subnetwork) and task-specific classification subnetworks for polyp and artifact detection.}
    \label{fig:lwf}
\end{figure*}

Following the LwF method, we first used our base artifact detector on all the images $X \in D^p$ to generate a set of artifact boxes. Then, we incorporated these predictions into $D^p$ as ground truth for artifacts. Setting the threshold at which we select these artifact annotations is an important hyperparameter. We thereby obtained a new dataset $D^{mt}$ that is suitable for training models on both tasks. The procedure is illustrated in Fig \ref{fig:lwf}a. 

We were then able to use $D^{mt}$ to train a MTL model $[Y^{t}, C^t] = f(X; \theta^s, \theta^t)$ with $t \in \{p, a\}$ using the loss function below. For simplicity, we set $Z^t = [Y^{t}, C^t]$:
\begin{equation} \label{eq:lwfl}
\mathcal{L} =\ell^p(Z^p, f(X; \theta^s, \theta^p)) + \ell^a(Z^a; f(X, \theta^s, \theta^a)) + R
\end{equation}
where $R=r(\theta^s, \theta^p, \theta^a)$ is a weight regularizer and $\ell^p$, $\ell^a$ are task-specific losses. In \cite{8107520} the two tasks have different loss functions, where the loss function $l^a$ for the initial task is the Knowledge Distillation loss.  The purpose of that loss is to encourage two networks to have the same output \cite{hinton2015distilling}. Given that our goal does not include maintaining good performance on the related task, we opted for using the same loss function, focal loss, for both the initial and the new task.

\subsection{MTL Models}  \label{subsec:mcl}

The LwF method requires having task-specific parameters for the polyp and artifact detection tasks. The RetinaNet architecture can naturally be extended to have a set of shared and task-specific parameters. We selected the parameters of the backbone and FPN to be our shared parameters $\theta^s$. Then, for our tasks $a$ and $p$, we built individual classification subnetworks with parameters $\theta^a$ and $\theta^p$, respectively. For the the regression subnetwork, parameters are shared across the tasks. Each of the three subnetworks has their own loss function. The framework is illustrated in  Figure \ref{fig:lwf}c.

\section{Experiments and Results} \label{sec:experiments_results}
    Our experiments are divided into two sections. First, section \ref{subsec:subsec_art_inf} evaluates how artifacts affect polyp detection performance. For this analysis, we used an in-house dataset that consists of 55,411 frames. The dataset is thereby far bigger than existing publicly available datasets and enabled us to get more statistically meaningful insights. In addition, existing publicly available datasets are often designed for polyp detection only, and therefore might remove frames that are strongly or entirely corrupted by artiacts. In contrast, our in-house dataset has not been processed at all and is thereby much closer to a real-life clinical setting. Section \ref{subsec:mtl_exp} then tests ways to make use of artifact knowledge to improve polyp detection. For this section, we conducted our results on publicly available datasets. We thus used different datasets for section \ref{subsec:subsec_art_inf} and  \ref{subsec:mtl_exp}. Thereby, we avoid bias by not using insights gained from the validation sets to design our multi-task approaches. Further, this will enable future work to compare their performance to us.
 
 \subsection{Datasets} \label{subsec:datasets}

\begin{figure}
\begin{center}
\includegraphics[width=0.99\columnwidth]{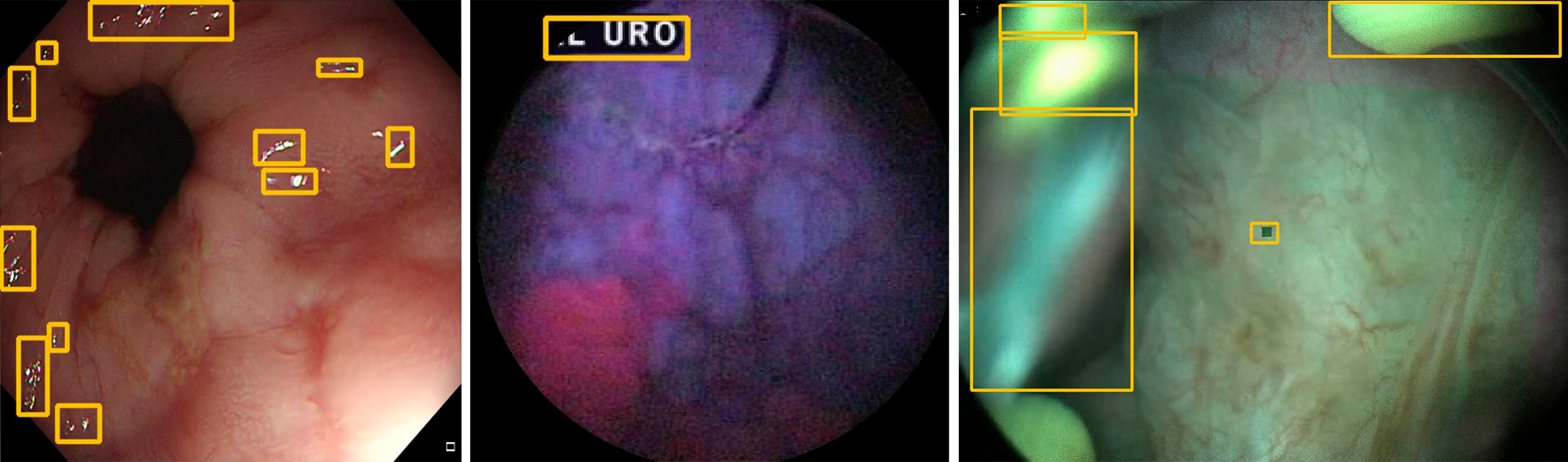}
\end{center}
\caption{a) Sample frames from the EAD challenge dataset, which show artifacts of the class miscellaneous (misc).} \label{fig:misc_sample}
\end{figure}

We make use of the following datasets:
\begin{itemize}

	\item \textbf{EAD2019} \cite{DBLP:journals/corr/abs-1905-03209}: The training data of the EAD2019 dataset contains 2192 unique video frames with bounding box annotations and class labels for seven artifact classes. The data displays a lot of variation, as it was obtained from four different centres and includes still frames from multiple tissues, light modalities, and populations. Example artifact annotations can be found in \ref{fig:cvc_samples} and \ref{fig:misc_sample}.

    \item \textbf{CVC-ClinicDB} \cite{BERNAL201599, bernal2017comparative}: The CVC-ClinicDB dataset is a publicly available colonoscopy dataset that consists of 612 colonoscopy frames taken from a total of 29 video sequences. The sequences are from routine colonoscopies and were selected to represent as much variation in polyp appearance as possible. The whole dataset contains 31 polyps and none of the images contain no polyps. Of these 31 polyps, 22 are small polyps below the size of 10mm and 9 are greater than 10mm in diameter \cite{fernandez2016exploring}. The colonoscopies were conducted with standard resolution white-light video colonoscope (Q160AL or Q165L). All frames have a pixel resolution of $388\times 284$ pixels in standard definition.
    
    \item \textbf{ETIS-Larib} \cite{bernal2017comparative}: ETIS-Larib is a polyp dataset that was used as the testing dataset in the still frame analysis task in \cite{bernal2017comparative}. The dataset consists of 196 high definition frames that were selected from 34 sequences. The dimension of the images is $1225\times 966$. In total the dataset contains 44 different polyps and all frames contain at least one polyp. The colonoscopy procedures were conducted with a Pentax 90i series device.

	\item \textbf{in-house}: Our in-house dataset was collected in the endoscopy department of Klinikum rechts der Isar, Technical University of Munich. It consists of 55,411 frames of size $1920 \times 1080$, obtained from 431 endoscopic videos. All frames were annotated semi- automatically and the annotations were validated by medical students. There is only one polyp ground-truth-box per frame.
	
	\item \textbf{Kvasir-SEG} \cite{jha2020kvasir}: The Kvasir-SEG dataset contains 1,000 polyp images with bounding boxes and segmentation masks. Image resolution varies between $332 \times 487$ and $1920\times 1072$ pixels. 
	
\end{itemize}{}

The EAD2019 dataset was used to train our artefact detection model. Our polyp detection model trained on the CVC-Clinic dataset. The in-house dataset was used to conduct our analysis on how artifacts affect the polyp detection model. Our multi-task approaches were evaluated on CVC-Clinic (3-fold cross-validation), ETIS-Larib, and Kvasir-SEG.

 \subsection{Polyp Detection Implementation Details}

Due to memory constrains, we used a batch size of 2 during training on polyp detection. As focal loss parameters, we found $\gamma=2.5$ and $\alpha=0.25$ to be optimal. As a backbone network, we chose ResNet-50 pre-trained on ImageNet. Despite achiever higher performance with ResNet-101 and ResNet-152, we chose ResNet-50 as the goal was to have a simple baseline as benchmark against our multi-task approaches. Except if stated otherwise, all our experiments with RetinaNet were trained with the Adam optimizer \cite{kingma2014adam} with a learning rate of $10e-5$ with decay by a factor of 10 when the improvement changes are minor between epochs. Basic data augmentation was applied. I consisted of a randomized combination of rotation, translation, shear, scaling, and flipping. Each image was rotated, translated, and sheared by a factor of -0.1 to 0.1, scaled between 0.9 to 1.1 of its original size, and flipped both horizontally and vertically at a probability of 50\%.

\subsection{Polyp Detection Metrics} \label{subsec:metrics}

In order to compare our polyp detection performance with existing methods, we follow the commonly used validation framework from \cite{bernal2017comparative}, where scores are reported by giving the number of true positives, false positives, false negatives, precision, recall, F1-score, and F2-score. This validation framework can be used both for segmentation and object detection approaches. A true positive is any detection where the centroid of the bounding box (or segmentation mask) lies within the polyp ground-truth mask. False positives are detections where the centroid is outside of the polyp mask. Each ground-truth can only have one true positive, thus each detection that correctly detects a polyp, which has already been detected, is a false positive. False negatives are all ground-truth polyps that have not been detected.

\subsection{Effects of Artifacts on Polyp Detection} \label{subsec:subsec_art_inf}

\begin{figure} 
    \centering
    \includegraphics[width=0.95\columnwidth]{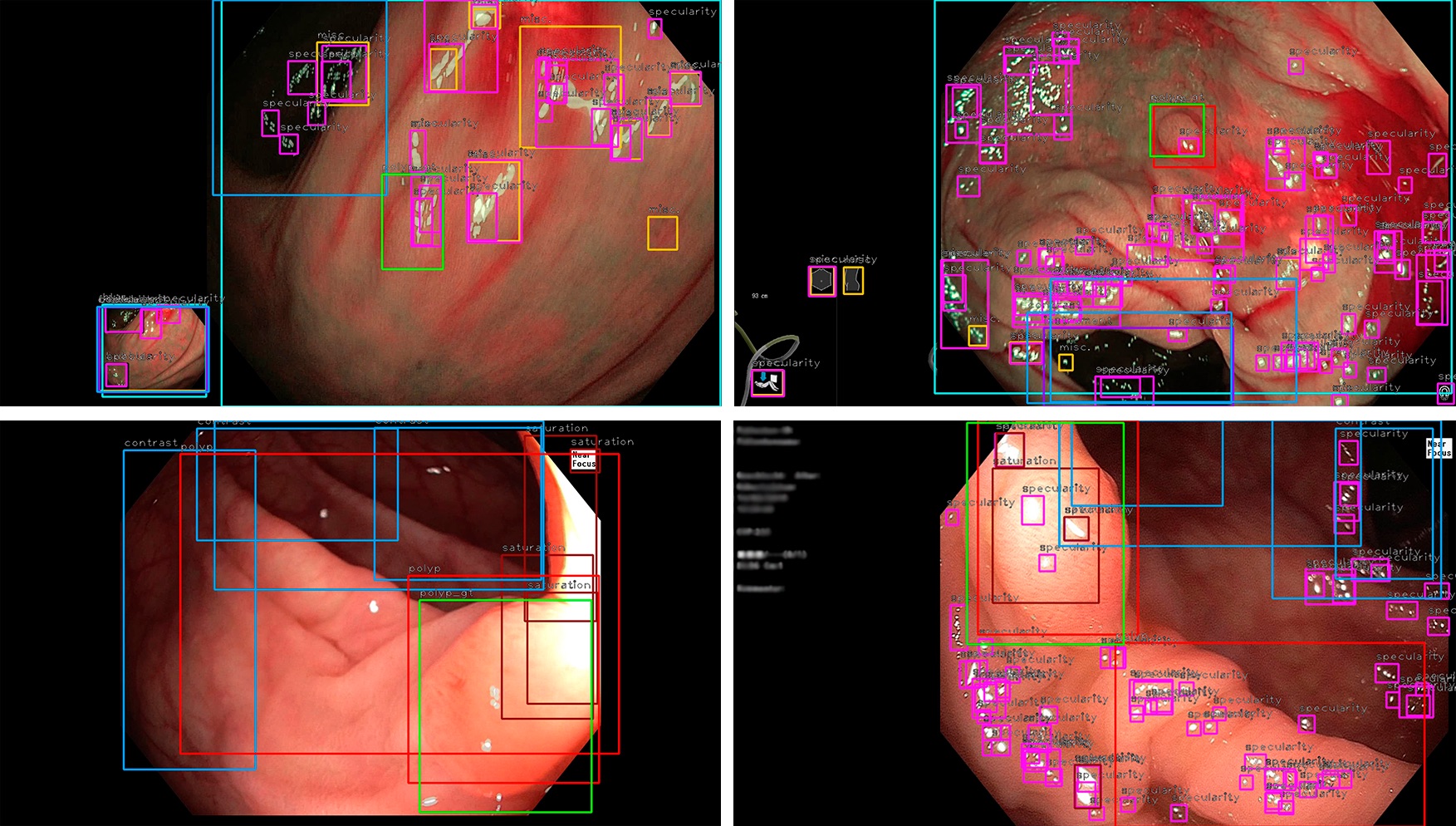}
    \caption{Examples of artifacts and polyps predictions in our in-house dataset. Green bounding boxes represent polyps ground truth and red bounding boxes are polyp predictions. The remaining bounding boxes are bubbles (black), specularity (pink), blur (blue), and saturation (brown). Best viewed in color.}
    \label{fig:mri_samples}
\end{figure}

The first contribution of this work is to investigate how deep learning based polyp detection performance is affected by the presence of image artifacts. We conducted these experiments on our in-house dataset. First, we used our artefact detection model to annotate artifact bounding boxes in that dataset. Artifact bounding boxes were taken at a confidence threshold of 0.25. We chose this value as we wanted to get the most possible information on artifacts and this threshold has proven optimal in our submissions to the EAD challenge. We then ran our polyp detection model to get a set of polyp bounding box predictions on the same images. Polyp detections are always considered at a threshold of 0.5, which has proven optimal in preliminary experiments. We then performed a number of analyses, as described below, to better understand how artifacts affect our polyp detector. Illustrative example frames can be found in Figure \ref{fig:mri_samples}.

\subsubsection{Artifact Presence vs. Performance} \label{artvperf}

\begin{figure}
    \centering
    \includegraphics[width=0.98\columnwidth]{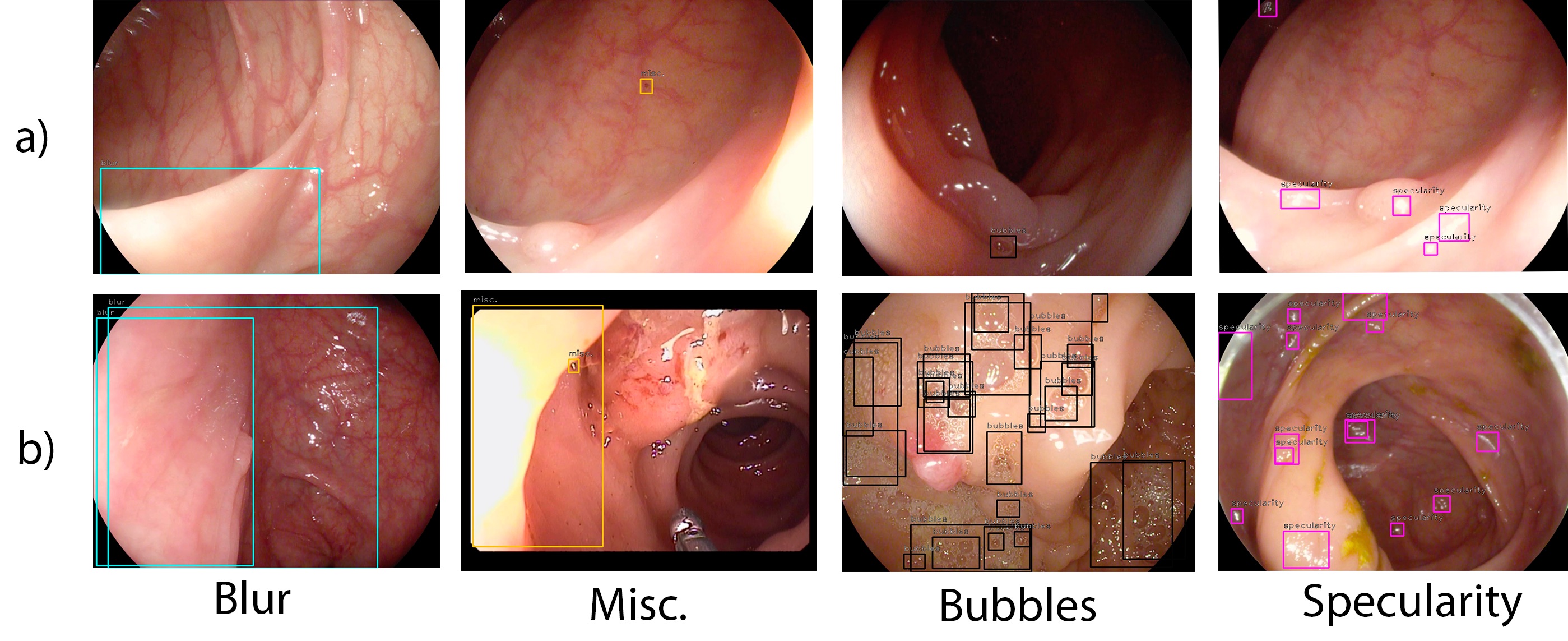}
    \caption{In experiment \ref{artvperf} we set area thresholds for some artifacts to consider them present in an image. The first row (a) shows artifacts that do not meet the are threshold, compared to the row below (b) where they meet the area threshold.}
    \label{fig:art_area_th}
\end{figure}

In our first analysis, we wanted to get a general idea of how polyp detection performance differs given the presence of artifacts. For each artifact class, we took the subset of images where the artifact is contained (at a given area threshold) and compared polyp detection performance to the subset of images where the artifact is not present. Table \ref{tab:pres} shows our results. To consider that an artifact is present in the image, we defined different area thresholds for the different artifacts (see Fig. \ref{fig:art_area_th}). The area was computed based on the total amount of pixels covered by a given artifact. For blur, the threshold was set at 50\% of the entire image size, to only include images where the entire image is blurred. For specularity, which is effectively present in all (99\%) of the images, we selected a threshold of 5\%, to cover only images where there is a high amount of specularity. For this threshold, specularity is still present in 56\% of images. For similar reasons, we selected 2\% of the image area as a threshold for misc. artifacts and bubbles.  No thresholds were set for contrast and saturation. 

\begin{table}[ht]
  \centering
  \caption{For six different artifact classes, polyp detection performance is compared between images where the artifact is present and images where the artifact is not present. Frequency gives the share of images where the artifact is present. The score difference is the difference of the respective metrics between images where the artifact is present or not present. The experiment was conducted of the in-house dataset, consisting of 55,411 frames.}
  \resizebox{\columnwidth}{!}{
    \begin{tabular}{l | c |c |c |c |c} 
    \hline
          &       & \multicolumn{4}{c}{Score Difference (\%)} \\
          \cline{3-6}
    Artifact type & Frequency (\%)  & precision & recall & F1    & F2 \\
    \hline 
    bubbles	& 9.51	&	-5.67	&	2.33	&	-1.61	&	0.76 \\
    blur	& 20.85	&	4.31	&	-10.05	&	-3.69	&	-7.67 \\
    misc.	& 48.99	&	5.62	&	-1.94	&	1.62	&	-0.56 \\
    specularity	& 49.28	&	-6.93	&	4.65	&	-0.80	&	2.54 \\
    saturation	& 69.28	&	0.20	&	2.54	&	1.44	&	2.12 \\
    contrast	& 79.70	&	-2.86	&	4.41	&	1.09	&	3.15 \\
    \hline
    \end{tabular}
    }
  \label{tab:pres}
\end{table}

\subsubsection{Artifacts Overlapping Polyp Detection} \label{subsubsec:overlap}

\begin{figure}
    \centering
    \includegraphics[width=0.9\columnwidth]{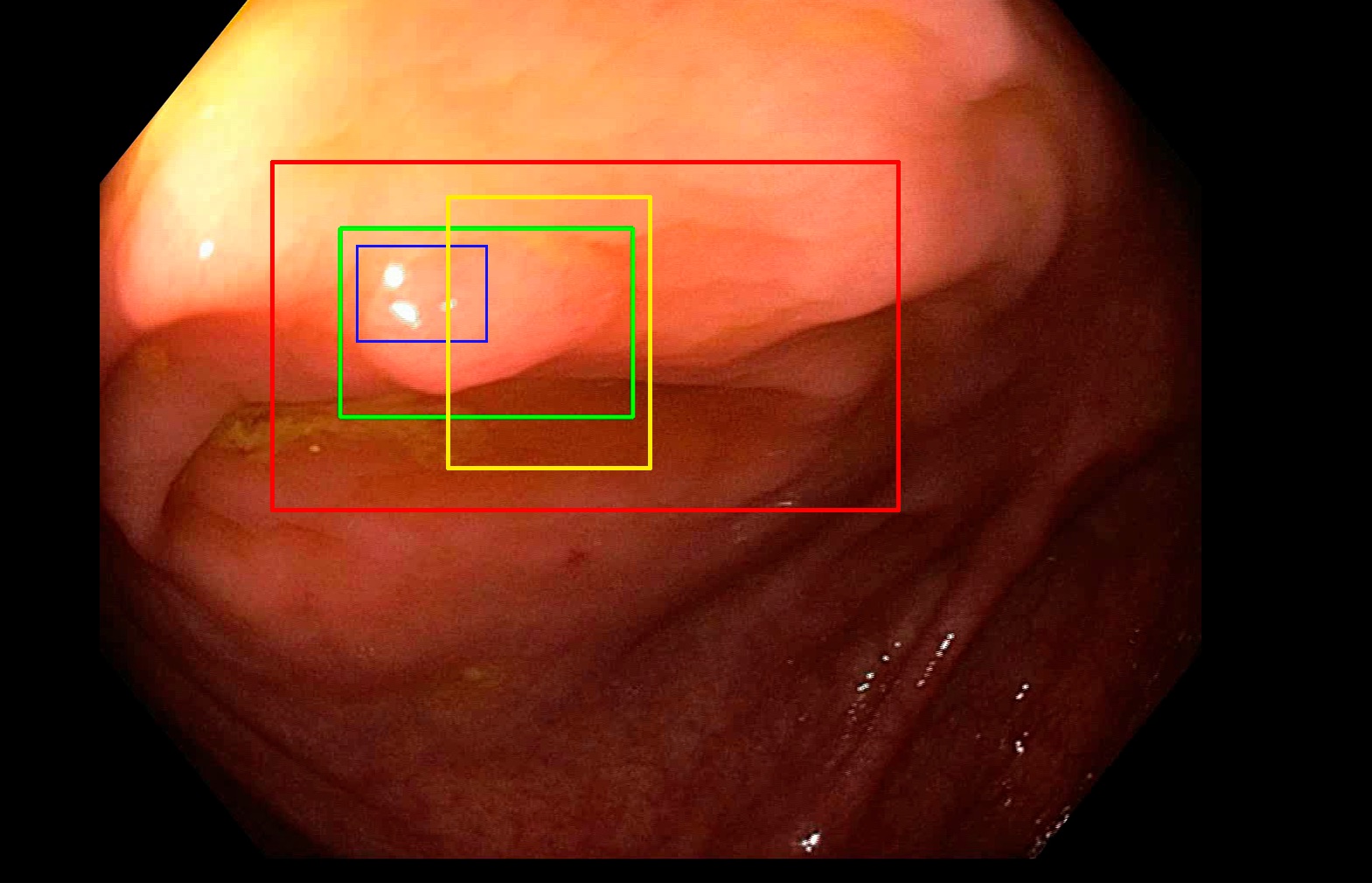}
    \caption{The green box represents a polyp detection. Artifact bounding boxes either overlap with a polyp (yellow) or are contained within a polyp (blue).}
    \label{fig:art_overlap}
\end{figure}

We next evaluated the frequency of overlap between artifacts detections and polyp ground-truths as well as true positive, false positive, and false negative polyp predictions.  We counted how many times these polyp bounding boxes overlap with an artifact (results in Table \ref{tab:ov}) and how many times a given artifact is inside of them  (results in Table \ref{tab:in}). See Fig. \ref{fig:art_overlap} for an example of these relationships. For the sake of this analysis, we did not remove a prediction if it overlapped an existing prediction. I.e. if two polyp detections overlap the same polyps, they are still both considered correct predictions (this was naturally not done in experiments where we evaluated polyp detection performance). Therefore, the sum of true positives and false positives does not add to the amount of ground-truth polyps in our results tables. The goal of this analysis was to understand how artifacts are affecting polyp detection at a more in-depth level. This analysis for example allowed us to find out whether a certain artifact is sometimes misclassified as a polyp (i.e. when false positive detection often overlap with that artifact). We considered an artifact and a polyo to overlap if their intersection-over-union (IoU) is greater than 0.5. 

\begin{table*}[ht]
  \centering
  \caption{The share of ground-truth polyps as well as true positive, false positive, and false negative polyp detections that overlap with respective artifacts (i.e. their intersection-over-union is greater than 0.5). Frequency gives the count of the different polyp bounding box types in the dataset. Any artifact incorporates all six artifacts. These experiments were conducted on our in-house dataset, which consists of 55,411 frames.}
    \resizebox{\textwidth}{!}{
    \begin{tabular}{l |c |c |c |c |c |c |c |c}
    \hline
          &       & \multicolumn{7}{c}{Share of polyps \textbf{overlapping} artifacts (\%)} \\
        \cline{3-9}   
    Polyp type & Frequency & Any artifact & Bubbles & Blur  & Misc. artifact & Specularity & Saturation & Contrast \\
    \hline
    ground-truth & 55411   & 17.8 & 1.5 & 3.4  & 2.2 & 2.7 & 4.5 & 2.9 \\
    true positives & 45532   & 17.4 & 1.3 & 3.5  & 1.6 & 2.2 & 3.8 & 3.0 \\
    false positives & 10435 & 23.8 & 3.0 & 6.1  & 2.7 & 1.9 & 1.3 & 5.8 \\
    false negatives & 13438    & 13.7 & 1.2 & 3.0  & 3.2 & 2.2 & 4.3 & 2.6 \\
    \hline
    \end{tabular}
    }
  \label{tab:ov}
\end{table*}

\begin{table*}[ht]
  \centering
  \caption{The share of ground-truth polyps as well as true positive, false positive, and false negative polyp detections that contain respective artifacts inside of them (i.e. the bounding box of the artifact is fully contained inside the polyp bounding box). Frequency gives the count of the different polyp bounding box types in the dataset. Any artifact incorporates all six artifacts. These experiments were conducted on our in-house dataset, which consists of 55,411 frames.}
    \resizebox{\textwidth}{!}{
    \begin{tabular}{l |c| c| c| c| c| c| c| c}
    \hline
          &       & \multicolumn{7}{c}{Share of polyps \textbf{containing} artifacts (\%)} \\
\cline{3-9}    Polyp type & Frequency & Any artifact & Bubbles & Blur  & Misc. artifact & Specularity & Saturation & Contrast \\
    \hline
    ground-truth & 55411&	88.5 & 11.0 & 0.3  & 33.1 & 86.8 & 12.7 & 0.7 \\
    true positives & 45532 &	94.0 & 15.4 & 0.5  & 37.3 & 91.9 & 16.2 & 2.5 \\
    false positives & 10435 &	92.0 & 22.7 & 1.7  & 37.8 & 87.9 & 16.7 & 6.1 \\
    false negatives & 13438 &	79.9 & 8.3 & 0.2  & 32.8 & 77.5 & 7.2 & 0.4 \\
    \hline
    \end{tabular}
    }
  \label{tab:in}
\end{table*}

\subsection{Multi-task Learning} \label{subsec:mtl_exp}

Next, we investigated approaches to incorporate artifact knowledge in polyp detection. We compared different such approaches, including our MTL model described in section \ref{subsec:mcl}. The results shown in this section are the averaged scores of  3-fold cross-validation on the CVC-ClinicDB dataset and follow the validation framework laid out in section \ref{subsec:metrics}. Our MTL approaches are enumerated from simple to more complex: we tested using the artefact model to initialize the weights of the polyp detector, we then considered it a multi-class problem (with different class weights), and finally we employed a specialized architecture to have a model with shared and task-specific parameters.

\subsubsection{Transfer Learning} \label{subsubsec:tl_exp}

In our first approach, we looked at the most straight-forward way to utilize the artifact data: using the weights of the artifact model for initializing the polyp detection model. This is thus a transfer learning approach, where we pre-trained our polyp detector on the EAD2019 artifact dataset and then fine-tuned it on the polyp dataset. We have tried two different TL approaches: 1) fine-tuning the entire network and 2) freezing the backbone and only fine-tuning the classification subnetwork of RetinaNet. We also compared the performance with models pre-trained on ImageNet and COCO. Results are shown in Table \ref{tab:tl}.

\begin{table}[ht]
  \centering
  \caption{Polyp detection performance of different Transfer Learning (TL) approaches. For "initialization", the entire model is fine-tuned on the final task. For "freeze backbone", only the final layers of the model are trained on the final task. "Baseline" is our polyp-only model that was pre-trained on ImageNet. Results are from 3-fold cross-validation on CVC-ClinicDB.}
  \resizebox{\columnwidth}{!}{
    \begin{tabular}{l |l| c| c| c| c}
    \hline
    Pre-training & TL Type & Precision & Recall & F1    & F2 \\
    \hline
    ImageNet & initialization & \textbf{0.849} & 0.814 & 0.829 & 0.820 \\
    COCO  & initialization & 0.833 & \textbf{0.862} & \textbf{0.845} & \textbf{0.855} \\
    Artifact Model & initialization & 0.832 & 0.859 & 0.845 & 0.853 \\
    Artifact Model & freeze backbone & 0.836 & 0.766 & 0.794 & 0.776 \\
    \hline
    \end{tabular}
    }
  \label{tab:tl}
\end{table}

\subsubsection{Multi-class Learning and  Artifact Confidence Threshold} \label{susubbsec:mcl_exp}

A naive approach for MTL on both polyps and artifacts is simply formulating the problem as a multi-class classification problem, where we have one class for each artifact and one class for the polyps. We thus used our single-task polyp detector, but extended it by six classes to include artifacts. We get our artifact annotations as described in section \ref{subsec:lwf}. We experimented with different thresholds for obtaining our artifact annotations. Results are shown in Table \ref{tab:st}.

\begin{table}[htbp]
  \centering
  \caption{Polyp detection performance of a multi-class RetinaNet model that detects polyps as well as artifacts. Artifact threshold gives the threshold for which we considered artifact annotations as ground-truth in the training set. The lower that threshold is, the more artifacts are contained in each image. "Baseline" is our polyp-only model that was pre-trained on ImageNet. Results are from 3-fold cross-validation on CVC-ClinicDB.}
  \resizebox{\columnwidth}{!}{
    \begin{tabular}{l| c| c| c| c| c}
    \hline
    Artifact & Artifacts per & Precision & Recall & F1    & F2 \\
    Threshold & Image & & & & \\
    \hline    
    0.2    & 37.8  & 0.146 & 0.515 & 0.225 & 0.336 \\
    0.4    & 7.7   & 0.237 & 0.487 & 0.310 & 0.392 \\
    0.5    & 3.6   & 0.413 & 0.690 & 0.516 & 0.608 \\
    0.6    & 2.0   & 0.722 & 0.640 & 0.673 & 0.652 \\
    0.8    & 1.2   & \textbf{0.927} & 0.721 & 0.810 & 0.754 \\
    \hline
    {baseline} & N.A.   & 0.849 & \textbf{0.814} & \textbf{0.829} & \textbf{0.820} \\
    \hline
    \end{tabular}
    }
  \label{tab:st}
\end{table}

\subsubsection{Weighting Artifact and Polyps Classes} \label{subsubsec:mclw_exp}

We repeated the previous experiments but increased the weighting of the polyp class by 25\%, 50\%, and 75\% of the classification loss, while distributing the remaining share of the weight equally between all six artifact classes. We further compared pre-training on ImageNet and the EAD2019 artifact dataset. In order to include rich artifact information, we have selected artifact annotations at a threshold of 0.5. For the EAD competition, optimal artefact thresholds were between 0.25 and 0.5. Results are displayed in Table \ref{tab:cw}.

\begin{table}
  \centering
  \caption{Polyp detection performance of a multi-class RetinaNet model that detects polyps as well as artifacts and has specific class weights. Proportional class weights of the polyp class are given in the first column. The remaining weight is distributed equally among the remaining artifact classes. Pre-training is given either from ImageNet or from our artifact detector. Artifacts in the training data are taken at a 0.5 confidence threshold. "Baseline" is our polyp-only model that was pre-trained on ImageNet. Results are from 3-fold cross-validation on CVC-ClinicDB.}
  \resizebox{\columnwidth}{!}{
    \begin{tabular}{l| l| c| c| c| c}
    \hline
    Polyp Weight & Pre-training & Precision & Recall & F1    & F2 \\
    \hline
    25\%  & ImageNet & 0.774 & 0.726 & 0.749 & 0.735 \\
    50\%  & ImageNet & 0.797 & 0.761 & 0.775 & 0.766 \\
    75\%  & ImageNet & 0.752 & 0.765 & 0.744 & 0.752 \\
    25\%  & Artifact Model & 0.808 & 0.766 & 0.785 & 0.773 \\
    50\%  & Artifact Model & 0.777 & 0.789 & 0.782 & 0.786 \\
    75\%  & Artifact Model & 0.770 & \textbf{0.825} & 0.796 & 0.813 \\
    \hline
    no weighting & ImageNet & 0.413 & 0.690 & 0.516 & 0.608 \\
    \hline
    baseline & ImageNet & \textbf{0.849} & 0.814 & \textbf{0.829} & \textbf{0.820} \\
    \hline
    \end{tabular}
    }
  \label{tab:cw}
\end{table}

\subsubsection{Double Classification Subnetwork} \label{subsubsec:double_subnetwork}

We then conducted experiments on our LwF inspired MTL model, which was described in section \ref{subsec:mcl}. We tried out four different weighting configurations for the loss function of the model. We weighted regression and artifact classification subnetworks by 1 and assigned weights of 1, 3, 10, and 20 to the polyp classification loss. We have tried training the models on artifacts taken at a 0.2 and at a 0.5 threshold. Results are shown in Table \ref{tab:ds}.

\begin{table}[htbp]
  \centering
  \caption{Polyp detection performance of our learning without forgetting (LwF) inspired multi-task model. The model has three subnetworks on top of the model backbone: a regression (reg) subnetwork, an artifact (art) classification subnetwork, and a polyp (pol) classification subnetwork. These three subnetworks have their own losses and the respective weights of these losses are given in the first column. The artifact thresholds for the training data are given in the second column. "Baseline" is our polyp-only model that was pre-trained on ImageNet. Results are from 3-fold cross-validation on CVC-ClinicDB.}
    \resizebox{\columnwidth}{!}{
    \begin{tabular}{c| c| c| c| c| c}
    \hline
    Loss Weights & Artifact & Precision & Recall & F1    & F2 \\
    (reg:art:pol) & Threshold & & & & \\
    \hline
    1:1:1 & 0.2 & 0.836 & 0.269 & 0.381 & 0.304 \\
    1:1:3 & 0.2 & 0.841 & 0.484 & 0.604 & 0.525 \\
    1:1:10 & 0.2 & 0.803 & 0.650 & 0.715 & 0.674 \\
    1:1:20 & 0.2 & 0.831 & 0.677 & 0.739 & 0.699 \\
    \hline
    1:1:1 & 0.5 & 0.820 & 0.727 & 0.771 & 0.744 \\
    1:1:3 & 0.5 & 0.802 & 0.755 & 0.773 & 0.761 \\
    1:1:10 & 0.5 & 0.779 & 0.793 & 0.781 & 0.787 \\
    1:1:20 & 0.5 & 0.796 & \textbf{0.818} & 0.804 & 0.812 \\
       \hline
    \multicolumn{2}{c |}{baseline} & \textbf{0.849} & 0.814 & \textbf{0.829} & \textbf{0.820} \\
    \hline
    \end{tabular}
    }
  \label{tab:ds}
\end{table}

\subsubsection{Removing artifact Classes} \label{subsubsec:red}

Section \ref{subsec:subsec_art_inf} has shown that some artifacts classes are more relevant to polyp detection performance than others. In addition, our models may be overwhelmed by the high number of artifact classes and reducing that number may be beneficial. We therefore restricted ourselves to artifact classes that appear most relevant for polyp detection. For instance, if we know that blurs are often misclassified as polyps, or that polyps containing specularity are easier to detect, we know that these classes can be useful for polyp detection. We created four different subsets of artifacts to include in our MTL approach. The first subset contains only the artifact that we deemed most important, the second set contains the two most prioritized ones, and so on. The artifacts that we deemed the most influential, in descending order, are the following: blur, specularity, misc. artifacts, and bubbles. We run our MTL model with artifacts thresholds at 0.5. The weights of the loss functions were chosen to be proportionate to the amount of artifacts in the image for the different artifact classes. For the four models in Table \ref{tab:dc}, they are given by 1:5:1, 1:1:3, 1:1:3, 1:1:3, respectively. Weighting corresponds to the regression, artifact, and polyp loss functions.

\begin{table}
  \centering
  \caption{Polyp detection performance of our learning without forgetting (LwF) inspired multi-task model when only incorporating knowledge about some of the existing artifact classes. The star (*) in the artifact column means that the given artifact was included for this model. Artifacts in the training data are taken at a 0.5 confidence threshold. "Baseline" is our polyp-only model that was pre-trained on ImageNet. Results are from 3-fold cross-validation on CVC-ClinicDB.}
    \resizebox{\columnwidth}{!}{
    \begin{tabular}{c| c| c| c| c| c| c| c}
    \hline
    \multicolumn{4}{c|}{Artifacts} & \multirow{2}{*}{Precision} & \multirow{2}{*}{Recall} & \multirow{2}{*}{F1} & \multirow{2}{*}{F2} \\
\cline{1-4}    blur  & spec. & misc. & bubbles    &       &       &       &  \\
    \hline
    *     &       &       &          & 0.870 & 0.808 & \textbf{0.836} & 0.819 \\
    *     & *     &       &          & 0.828 & \textbf{0.826} & 0.825 & \textbf{0.825} \\
    *     & *     & *     &          & 0.843 & 0.820 & 0.829 & 0.823 \\
    *     & *     & *     & *        & 0.859 & 0.790 & 0.821 & 0.802 \\
    \hline
    \multicolumn{4}{c|}{baseline}  & 0.849 & 0.814 & 0.829 & 0.820 \\
    \hline
    \end{tabular}
    }
  \label{tab:dc}
\end{table}

\subsection{Understanding the Effects}

Finally, we repeated some experiments from subsection \ref{subsubsec:overlap} in order to get an understanding of how leveraging artifact information has changed the way that artifacts affect polyp detection performance. We selected the MTL RetinaNet trained on blur, bubbles, misc., and specularity. Artifact annotations for training were taken at a 0.2 threshold, since it is similar to the artifact threshold of 0.25 taken in the experiments in \ref{subsubsec:overlap}. Results are shown in Table \ref{tab:dc_ov} and \ref{tab:dc_in}.

\begin{table*}[ht]
  \centering
  \caption{The share of ground-truth polyps as well as true positive, false positive, and false negative polyp detections that overlap with respective artifacts (i.e. their intersection-over-union is greater than 0.5). For each of the four artifacts, we compare how the predictions of the baseline model are affected vs. how the predictions of our learning without forgetting (LwF) inspired multi-task model are affect. The multi-task model incorporates knowledge from the four artifacts in this table. Frequency gives the count of the different polyp bounding box types in the dataset. These experiments were conducted on the ETIS-Larib dataset.}
    \begin{tabular}{l| cc|cc|cc|cc|cc}
    \hline
          &       & \multicolumn{1}{c|}{} & \multicolumn{8}{c}{Share of polyps \textbf{overlapping} artifacts (\%)} \\
\cline{4-11}          & \multicolumn{2}{c|}{frequency} & \multicolumn{2}{c}{bubbles} & \multicolumn{2}{c}{blur} & \multicolumn{2}{c}{misc. artifact} & \multicolumn{2}{c}{specularity} \\
\cline{2-11}    \multicolumn{1}{c|}{polyp type} & polyps & \multicolumn{1}{c|}{MTL} & polyps & \multicolumn{1}{c|}{MTL} & polyps & \multicolumn{1}{c|}{MTL} & polyps & \multicolumn{1}{c|}{MTL} & polyps & MTL \\
    \hline
    ground-truth & 208   & 208   & 3.4   & 3.4   & 1.0     & 1.0     & 0.5   & 0.5   & 4.8   & 4.8 \\
    true positives & 137   & 152   & 2.2   & 3.3   & 1.5   & 2.0     & 0.0     & 0.0     & 3.6   & 3.9 \\
    false positives & 39    & 64    & 5.1   & 6.2   & 30.8  & 6.2   & 0.0     & 3.1   & 5.1   & 4.7 \\
    false negatives & 63    & 64    & 6.3   & 4.7   & 1.6   & 0.0     & 1.6   & 1.6   & 6.3   & 6.2 \\
    \hline
    \end{tabular}
  \label{tab:dc_ov}
\end{table*}

\begin{table*}[ht]
  \centering
  \caption{The share of ground-truth polyps as well as true positive, false positive, and false negative polyp detections that contain respective artifacts inside of them (i.e. the bounding box of the artifact is fully contained inside the polyp bounding box). For each of the four artifacts, we compare how the predictions of the baseline model are affected vs. how the predictions of our learning without forgetting (LwF) inspired multi-task model are affect. The multi-task model incorporates knowledge from the four artifacts in this table. Frequency gives the count of the different polyp bounding box types in the dataset. These experiments were conducted on the ETIS-Larib dataset.}

    \begin{tabular}{l| cc|cc|cc|cc|cc}
    \hline
          &       & \multicolumn{1}{c|}{} & \multicolumn{8}{c}{Share of polyps \textbf{containing} artifacts (\%)} \\
\cline{4-11}          & \multicolumn{2}{c|}{frequency} & \multicolumn{2}{c}{bubbles} & \multicolumn{2}{c}{blur} & \multicolumn{2}{c}{misc. artifact} & \multicolumn{2}{c}{specularity} \\
\cline{2-11}    \multicolumn{1}{c|}{polyp type} & polyps & \multicolumn{1}{c|}{MTL} & polyps & \multicolumn{1}{c|}{MTL} & polyps & \multicolumn{1}{c|}{MTL} & polyps & \multicolumn{1}{c|}{MTL} & polyps & MTL \\
    \hline
    ground-truth & 208   & 208   & 14.4  & 14.4  & 0.0     & 0.0     & 7.2   & 7.2   & 60.6  & 60.6 \\
    true positives & 137   & 152   & 11.7  & 17.8  & 0.0     & 0.0     & 10.2  & 10.5  & 75.9  & 79.6 \\
    false positives & 39    & 64    & 23.1  & 15.6  & 15.4  & 1.6   & 25.6  & 17.2  & 79.5  & 64.1 \\
    false negatives & 63    & 64    & 19.0    & 7.8   & 0.0     & 0.0     & 6.3   & 3.1   & 41.3  & 28.1 \\
    \hline
    \end{tabular}
  \label{tab:dc_in}
\end{table*}

\subsection{Generalizability: testing our models on different datasets} \label{subsec:mtl_sota}

So far, all the results in section \ref{subsec:mtl_exp} were from a 3-fold cross validation on the CVC-ClinicDB dataset. To verify the validity of these results on other datasets, we re-run the best-performing model of each experiment in sections \ref{subsec:mtl_exp} on the ETIS-Larib dataset, our in-house dataset, and the Kvasir-SEG dataset. For this, all our models were trained on CVC-ClinicDB. While the aim of this work is not to surpass the state-of-the-art in polyp detection, thiss also enabled us to place the performance of our models in context of existing methods. The results are given in Table \ref{tab:mtl_sota}.

\begin{table}[htbp]
  \centering
  \caption{Polyp detection performance on three different datasets: ETIS-Larib (ETIS), our in-house dataset (in-house), and Kvasir-SEG (Kvasir). We test our best performing models of each of the experiments in section \ref{subsec:mtl_exp}. Our models were all trained on CVC-ClinicDB. If available, we compared our results to existing state-of-the-art methods that used the same evaluation framework. Note that the aim of this work is not to surpass the state-of-the-art. Only F1 scores are given.}
      \resizebox{\columnwidth}{!}{
    \begin{tabular}{l| c| c| c| c}
    \hline
    \multicolumn{1}{c|}{Method} & ETIS & in-house & Kvasir \\
    \hline
    State-of-the-art  & & & \\
    - \cite{bernal2017comparative}  & 0.708 & - & -\\
    - \cite{bernal2017comparative}  & 0.662 & - & -\\
    - \cite{shin2018automatic}  & \textbf{0.833} & - & -\\
    \hline
    Ours  & & & \\
    - baseline & \textbf{0.668} & \textbf{0.779} & \textbf{0.857}\\
    - Transfer learning (\ref{subsubsec:tl_exp}) & 0.621 & 0.747 & 0.816\\
    - Weighted multi-class (\ref{subsubsec:mclw_exp}) & 0.637 & 0.604 & 0.742\\
    - LwF (\ref{subsubsec:double_subnetwork}) & 0.618 & 0.650 & 0.766\\
    - LwF* (\ref{subsubsec:red}) & 0.642 & 0.695 & 0.846\\
    \hline
    \end{tabular}
    }
  \label{tab:mtl_sota}
\end{table}

\section{Discussion} \label{sec:discussion}
    \subsection{Effects of Artifacts on Polyp Detection} \label{disc:e1}

\paragraph{Overall correlation}

Table \ref{tab:pres} gives an initial indication that the presence of some artifacts affects the polyp detection rate. Indeed, artifacts like blur and bubbles affect the F1-score negatively (-3.69\% and -1.61\%). For bubbles, this is due to lower precision. This could be explained by the fact that bubbles intuitively look similar to polyps, and thus potentially be misclassified as such. For blur, lower F1-score is due to much lower recall (-10.05\%) on these images. At the same time, images containing misc., saturation, and contrast tend to have higher F1-scores. 

\paragraph{Causality vs. Correlation}

To get more insights into causality vs. correlation, we also looked at the correlation between the presence of different artifacts. We computed this by looking at how often pairs of artifact classes occur in the same image. This allowed us to better understand whether an artifact perhaps only affects performance because it correlates to the presence of another artifact, which actually causes it. Table \ref{tab:corr} shows that there is no meaningful correlation between any of the artifacts. Only misc. artifacts are slightly more correlated to others, which may be explained by the fact that they are loosely defined and their distribution may overlap with other classes. Another way to better understand causal effects is to look at the effect on performance given the location of an artifact with respect to the polyp. For instance, if we are able to concretely show that bubbles are likely to be misclassified as polyps, we make a stronger case for bubbles actually \emph{causing} a drop in performance (compared to e.g. simply being correlated to pictures where the polyp is very hard to detect). Nonetheless, further work is required to fully understand the causal effects of artifacts on the polyp detector (see \cite{castro2019causality}). 

\begin{table}[htbp]
  \centering
  \caption{The correlation between the presence of different artifacts on the in-house dataset.}
  \resizebox{\columnwidth}{!}{
    \begin{tabular}{l | c c c c c c}
    \hline
    & bubbles	&	blur	&	misc.	&	specularity	&	saturation	&	contrast	\\
    \hline
    bubbles	&	1.00	&	-0.12	&	-0.14	&	0.00	&	-0.07	&	-0.05	\\
    blur	&	-0.12	&	1.00	&	0.15	&	-0.02	&	0.04	&	-0.11	\\
    misc.	&	-0.14	&	0.15	&	1.00	&	-0.01	&	0.09	&	0.05	\\
    specularity	&	0.00	&	-0.02	&	-0.01	&	1.00	&	0.00	&	-0.01	\\
    saturation	&	-0.07	&	0.04	&	0.09	&	0.00	&	1.00	&	0.06	\\
    contrast	&	-0.05	&	-0.11	&	0.05	&	-0.01	&	0.06	&	1.00	\\
    \hline
    \end{tabular}
    }
  \label{tab:corr}
\end{table}

\paragraph{Artifacts and their location}

By assessing the correlation between artifacts and polyp detections and polyp ground-truths, Table \ref{tab:ov} gives a more in-depth view of how artifacts affect polyp detection. We observe that detections which are false positives more frequently overlap with artifacts (23.8\%) compared to actual ground-truth polyps (17.8\%), suggesting artifacts are frequently misclassified as polyps. Amongst others, these results seem to confirm our previous hypothesis that bubbles can be misclassified as polyps. Indeed, bubbles are twice as likely to overlap false positive polyp detections than ground-truth polyps. Albeit less intuitive, similar observations can be made on blur and contrast. At the same time, polyps that were missed by our polyp detector are less likely to overlap with an artifact than polyps overall (13.7\% vs. 17.8\%). This suggests that polyps that overlap with an artifact (or look like an artifact according to our artifact detector), are less likely to be missed. Saturated regions overlap with 4.5\% of ground-truth polyps and with 1.3\% of false positives. This indicates that well-lid regions (i.e. which are saturated) are less prone to being misclassified as polyps. While Table \ref{tab:in} shows that false positive polyp detections contain artifacts inside of them more often than ground-truth polyps do (92.0\% vs. 88.5\%), this is even more so the case for true positives (94.0\%). In addition, missed polyps (false negatives) contain artifacts less frequently (79.9\%) than the ground-truth. Therefore, this suggests that artifacts inside of a polyp region make polyps easier to detect and improve both precision and recall. This is especially true for specularity and saturation. For bubbles, blur, and contrast, we confirm previous results that indicate that they lead to more false positives.

In conclusion, Table \ref{tab:pres} gave us an initial idea of how are polyp detection performance is affected by artifact presence. It already indicated that artifacts, depending on their class, can either beneficiate or harm polyp detection. Tables \ref{tab:ov} and \ref{tab:in} gave us a more in-depth understanding on how different artifacts, at different locations with respect to the polyps, affect the polyp detector. Bubbles, blur, and contrast were found to lead to poorer performance. In contrast, saturation and specularity improved performance. Either way, this suggests that incorporating artifact knowledge into a polyp detection model could be beneficial. The hypothesis is that by being able to learn artifact representations, a model can learn to distinguish them from polyps and better leverage them to classify bounding boxes correctly. 

\subsection{Multi-task Learning}

\paragraph{Artifact annotator trustworthiness}

For training our multi-task approaches, we added artifact annotations to the CVC-ClinicDB dataset using an artifact detection model that was trained on the EAD2019 dataset. To verify that these artifact annotations are not entirely corrupt and thereby detrimental to our efforts, we randomly selected 60 images (i.e. around 10\% of the total) and manually inspected their artifact annotations. We inspected artifacts at a threshold of 0.25, which led to 1,422 annotations. While this did not give us insights on recall, we found a precision of 82.4\%. This let us conclude that the artifact detector is performing sufficiently well on CVC-ClinicDB and can be used for our multi-task approaches.

\paragraph{Overall performance}

We tested out different multi-task learning approaches to incorporate artifact knowledge, ranging from very simple to more complex, and compared them to a baseline that was only trained on a polyp dataset.  The first experiment (Table \ref{tab:tl}) shows that simply using the artifact dataset for pre-training yielded unsatisfactory results, with performance not surpassing pre-training on natural image datasets. Table \ref{tab:st} shows that modeling it as a simple multi-class problem, where the six artifacts and polyps constitute the seven classes, did not improve polyp detection performance either. Indeed, there seems to be an almost linear negative correlation between the confidence threshold of artifacts in the training data and the final polyp detection performance. In the original EAD artifact dataset, there are around 8.3 artifact annotations per image. If we suppose a similar distribution in the CVC-ClinicDB set, then an artifact threshold of 0.4, (which yields 7.7 artifacts per image) seems to be closest to reality. However, at this threshold, our model was performing very poorly (F1-score of 0.31), suggesting this method is very inadequate. Even when trying to improve the method by weighting the polyp class disproportionately (Table \ref{tab:cw}), performance increased but remained underwhelming. A possible explanation is that having eight different classes for a single classification subnetwork complicates the optimization. To address this, we built our double classification subnetwork MTL model, for which the results are in Table \ref{tab:ds}. On a NVIDIA TITAN Xp with 12 GB of RAM, this model takes around 0.06 seconds per image during inference. At an artifact threshold of 0.2, performance is poor regardless of the weighting configurations, suggesting that these artifacts are simply too noisy and too numerous. At a threshold of 0.5, we obtained decent performance even without weighting. However, despite improving over the multi-class approach, the model still underperformed the baseline. The discussion in section \ref{disc:e1} has shown that not all artifacts have an equal effect on polyp detection performance. We thus tried only incorporating knowledge on some artifacts in the MTL model. Table \ref{tab:dc} shows that this lead to performance at par or even slightly better than the polyp model. Nonetheless, performance is not significantly better and we have failed to show that incorporating artifact knowledge can indeed improve polyp detection rates.

\paragraph{Artifact robustness}

To better understand how the model behaviour changed after incorporating artifact knowledge, we re-ran some of the experiments from section \ref{subsubsec:overlap} with our MTL model from Table \ref{tab:dc} that included four artifact classes. Table \ref{tab:dc_ov} shows that e.g. for blur we have drastically reduced the number of blur artifacts that are misclassified as polyp. Table \ref{tab:dc_in} also shows that for all four artifacts, the MTL model has a lower share of false positives containing them, which suggests the model has been able to make use of artifact-related features in order to reduce the times it misclassified background regions that contain these artifacts as polyps. In addition, bubbles and specularity are contained less frequently in false negatives for the MTL model than for the polyp-only model (19\% vs. 7.8\% and 41.3\% vs. 28.1\%), which indicates that learning the features of these artifacts has helped the algorithm detect more polyps that are covered by them.

Our multi-task approaches have not yet made a convincing case for using artifact knowledge in polyp detection. However, experiments from Table \ref{tab:dc_ov} and \ref{tab:dc_in} show encouraging signs and warrant further research in this direction. Our multi-task approaches have been fairly naive and more sophisticated approaches might be able to reap the benefits of artifact knowledge. Besides incorporating artifact knowledge in a model by making it learn the artifact representation, other approaches could be considered, such as combining artifact and polyp detections to yield an uncertainty score for the model's polyp detections.

\section{Conclusion} \label{sec:conclusions}
    This work has contributed to a more thorough understanding of how endoscopic artifacts affect deep learning based polyp detection models. This was achieved by not only looking at artifact labels on an image level, as was done by \cite{bernal2017comparative}, but by having bounding box annotations, which specified the exact location of artifacts. The analysis has shown that some artifacts, such as bubbles or blur, may deteriorate polyp detection and others, such as specularity, can actually lead to better detection capabilities. We built upon this knowledge to extend a simple baseline polyp detector by making it simultaneously learn representation of artifacts and polyps. While we have not yet established significant improvements in performance by using this method, it still showed promising results as it managed to overcome some of the artifact-related challenges that automated polyp detectors face. Future work can build upon our analysis and multi-task approaches to ultimately improve polyp detection capabilities. 

\section*{Acknowledgments}
R. D. S. is supported by Consejo Nacional de Ciencia y Tecnolog\'{i}a (CONACYT), Mexico. 
S.A. is supported by the PRIME programme of the German Academic Exchange Service (DAAD) with funds from the German Federal Ministry of Education and Research (BMBF) 

\bibliographystyle{apalike}
\bibliography{references}

\end{document}